\pdfoutput=1
\documentclass[11pt]{article}

\usepackage{EMNLP2022}

\usepackage{times}
\usepackage{latexsym}

\usepackage[T1]{fontenc}
\usepackage[utf8]{inputenc}

\usepackage{microtype}

\usepackage{inconsolata}

\usepackage{caption}
\usepackage{subcaption}

\usepackage{xstring}
\usepackage{color, colortbl}
\usepackage{xcolor}
\usepackage{pgf}
\usepackage{collcell}
\usepackage{adjustbox}
\usepackage{multirow}
\usepackage{pdflscape}
\usepackage{booktabs}
\usepackage{float}

\usepackage{cleveref}
\crefformat{section}{\S#2#1#3} % see manual of cleveref, section 8.2.1
\crefformat{subsection}{\S#2#1#3}
\crefformat{subsubsection}{\S#2#1#3}

\usepackage[colorinlistoftodos]{todonotes}

\usepackage{philex}

\newcommand*{\wifce}{\textsc{w\&i-fce}}
\newcommand*{\wiked}{\textsc{wiked}}
\newcommand*{\wikedsampled}{\textsc{wiked-s}}
\newcommand*{\marvinlinzen}{\textsc{m\&l}}

\definecolor{darkred}{rgb}{0.0, 0.3, 0.6}
\definecolor{darkgreen}{rgb}{0.0, 0.4, 0.2}

\newcommand{\ApplyRedGradient}[1]{%
    \IfBeginWith{#1}{0.}{%
        \pgfmathsetmacro{\PercentColor}{60.0*#1}%
            \edef\x{\noexpand\cellcolor{darkred!\PercentColor}}\x\textcolor{black}{#1}%
    }%
    {%
        \IfBeginWith{#1}{1.}{%
            \pgfmathsetmacro{\PercentColor}{65.0*#1}%
                \edef\x{\noexpand\cellcolor{darkred!\PercentColor}}\x\textcolor{black}{#1}%
        }%
        {%
          \pgfmathsetmacro{\PercentColor}{0.001}%
          #1%
        }%
    }%
}
\newcommand{\ApplyGreenGradient}[1]{%
    \IfBeginWith{#1}{0.}{%
        \pgfmathsetmacro{\PercentColor}{60.0*#1}%
            \edef\x{\noexpand\cellcolor{darkgreen!\PercentColor}}\x\textcolor{black}{#1}%
    }%
    {%
        \IfBeginWith{#1}{1.}{%
            \pgfmathsetmacro{\PercentColor}{65.0*#1}%
                \edef\x{\noexpand\cellcolor{darkgreen!\PercentColor}}\x\textcolor{black}{#1}%
        }%
        {%
          \pgfmathsetmacro{\PercentColor}{0.001}%
          #1%
        }%
    }%
}
\newcommand{\ApplyGradientScaled}[1]{%
    \IfBeginWith{#1}{0.}{%
        \pgfmathsetmacro{\PercentColor}{(40.0*#1)}%
            \edef\x{\noexpand\cellcolor{darkred!\PercentColor}}\x\textcolor{black}{#1}%
    }%
    {%
      \pgfmathsetmacro{\PercentColor}{0.001}%
      #1%
    }%
}

\newcolumntype{R}{>{\collectcell\ApplyRedGradient}r<{\endcollectcell}}
\newcolumntype{G}{>{\collectcell\ApplyGreenGradient}r<{\endcollectcell}}

\title{Probing for targeted syntactic knowledge through grammatical error detection}

\author{
    {\bf Christopher Davis}\textsuperscript{$\dagger$} ~~~~
    {\bf Christopher Bryant}\textsuperscript{$\dagger$} ~~~~
    {\bf Andrew Caines}\textsuperscript{$\dagger$} \\
    {\bf Marek Rei}\textsuperscript{$\ddagger$$\dagger$} ~~~~
    {\bf Paula Buttery}\textsuperscript{$\dagger$} \\
    \textsuperscript{$\dagger$}ALTA Institute, Department of Computer Science \& Technology,\\ University of Cambridge, U.K. \\
    \textsuperscript{$\ddagger$}Imperial College London, U.K. \\
    \textsuperscript{$\dagger$}\texttt{\{ccd38,cjb255,apc38,pjb48\}@cam.ac.uk} \\
    \textsuperscript{$\ddagger$}\texttt{marek.rei@imperial.ac.uk}}

\begin{document}
\maketitle
\begin{abstract}

% There is some evidence that pre-trained language models encode syntactic information through targeted evaluation studies testing knowledge of subject-verb agreement (SVA).

Targeted studies testing knowledge of subject-verb agreement (SVA) indicate that pre-trained language models encode syntactic information. We assert that if models robustly encode subject-verb agreement, they should be able to identify when agreement is correct and when it is incorrect. To that end, we propose grammatical error detection as a diagnostic probe to evaluate token-level contextual representations for their knowledge of SVA. We evaluate contextual representations at each layer from five pre-trained English language models: \textsc{bert}, \textsc{xlnet}, \textsc{gpt-2}, \textsc{roberta}, and \textsc{electra}. We leverage public annotated training data from both English second language learners and Wikipedia edits, and report results on manually crafted stimuli for subject-verb agreement. We find that masked language models linearly encode information relevant to the detection of SVA errors, while the autoregressive models perform on par with our baseline. However, we also observe a divergence in performance when probes are trained on different training sets, and when they are evaluated on different syntactic constructions, suggesting the information pertaining to SVA error detection is not robustly encoded.

\end{abstract}

\section{Introduction}

Recent work investigates whether linguistic information is encoded in pre-trained transformer-based language models \cite{peters-etal-2018-deep, DBLP:conf/naacl/DevlinCLT19}. Research using diagnostic methods \cite{shi2016does, DBLP:conf/iclr/AlainB17, DBLP:conf/iclr/AdiKBLG17, conneau-etal-2018-cram, hupkes2018visualisation} indicates models encode syntax via experiments targeting, for example, part-of-speech and dependency labelling \cite{tenney2019you, jawahar:hal-02131630, hewitt-manning-2019-structural}, while targeted syntactic evaluation studies show models encode a large amount of hierarchical syntactic information in tests for subject-verb agreement \cite{linzen2016assessing, marvin-linzen-2018-targeted, goldberg2019assessing}. Although previous research has covered a large number of probing tasks \cite{tenney2019you, Liu2019-lg}, no one has yet fully explored grammatical error detection (GED) as a probe. We assert that the ability to detect ungrammatical tokens serves as a complementary evaluation to assess linguistic competence.

GED is a natural and complex NLP task that assesses a model's ability to detect which tokens in a sentence are grammatically incorrect. Ungrammatical tokens may be categorised within a taxonomy\footnote{Much recent research in GED uses error-type labels based on ERRANT \cite{bryant2017automatic}.} comprising three operational categories (replacement, unnecessary, and missing) and twenty-five categories based on parts-of-speech. For example:

\ex.\label{ex:r-verb-sva} [Replacement subject-verb agreement] \emph{The train \textcolor{red}{are} a good option for long trips.}

\ex.\label{ex:r-pron} [Replacement pronoun] \emph{Everybody must have free time for \textcolor{red}{yourself}.}

\ex.\label{ex:m-det} [Missing determiner] \emph{The birth of [\textcolor{red}{a}] new star.}

\ex.\label{ex:u-prep} [Unnecessary preposition] \emph{Public transport means travelling around [...] \textcolor{red}{by} using trains, buses, and planes.}

To do well in the task, a model must encode and make use of a wide array of linguistic information. For example, detecting subject-verb agreement errors in English tests a model's capacity to identify i) verbs, ii) the subjects of the verbs, iii) the grammatical number (singular/plural) of both, and iv) whether their number agrees.

The above makes the task a very interesting test-bed for evaluating a model's syntactic knowledge. We operationalise the GED task and train probes to detect replacement subject-verb agreement errors (as in Example \ref{ex:r-verb-sva}) using contextual representations from different hidden layers from five pre-trained English language models -- \textsc{bert} \cite{DBLP:conf/naacl/DevlinCLT19}, \textsc{xlnet} \cite{yang2019xlnet}, \textsc{gpt-2} \cite{radford2019language}, \textsc{roberta} \cite{liu2019roberta}, and \textsc{electra} \cite{clark-etal-2020-electra}.

To ensure a robust and thorough evaluation, we leverage existing publicly annotated data from two domains for training: essays by learners of English as a second language from both the Cambridge English Write \& Improve + LOCNESS (W\&I) corpus \citep{bryant2019bea} and the First Certificate in English corpus (FCE) \cite{yannakoudakis-etal-2011-new}, along with a corpus of automatically extracted edited sentences from native English Wikipedia edit histories \cite{wiked2014}. For evaluation, we re-frame the minimal-pair dataset from \citet{marvin-linzen-2018-targeted} to create targeted evaluation sets annotated for GED. In doing so, we demonstrate how existing minimal-pair datasets can be leveraged to create challenging and interpretable test sets for GED models\footnote{In principle these minimal-pair datasets can also be used to evaluate grammatical error correction systems.} \cite{hu-etal-2020-systematic}.

We find that \textsc{electra}, \textsc{bert}, and \textsc{roberta} linearly encode information for the detection of SVA errors in the contextual representations of verbs, however, we observe a gap in performance when probes are trained on data from different domains, implying the information is not encoded consistently or robustly. The results show consistent patterns across layers: both \textsc{bert} and \textsc{electra} encode information related to SVA errors in the middle-to-late layers, while \textsc{roberta} seems to encode information earlier in the model. Probes trained on representations from \textsc{gpt-2} and \textsc{xlnet} (with uni- and bi-directional decoding) perform poorly on the evaluation set, indicating a fundamental difference from either the training objective or pre-training data. Finally, we show that GED probes can complement existing tools for syntactic evaluation: our results suggest that although neural language models perform well on targeted syntactic evaluation tasks, their encoding of SVA does not robustly extend to the detection of SVA errors.\footnote{We release our code at \url{https://github.com/chrisdavis90/ged-syntax-probing}}

% Results show \textsc{bert} and \textsc{electra} encode the most useful information in the middle-to-late layers, while probes trained on \textsc{roberta} tend to perform best at layer 5 (out of 12). On the other hand, we find probes trained using representations from \textsc{gpt-2} perform poorly on the evaluation set, indicating that \textsc{gpt-2} does not linearly encode information in the representations of verbs relevant for SVA error detection. Finally, probes trained on \textsc{xlnet} don't perform much better, obtaining F$_{1}$ scores $\leq$0.6 with bidirectional decoding and $\leq$ 0.5 with sequential decoding.

\begin{table*}[ht]
% \small
  \centering
    \begin{tabular}{ll}
    \hline
    Syntactic Construction & Example \\
    \hline
    Simple agreement & The \textbf{author} \underline{laughs}/\underline{laugh$^{*}$} \\
    Agreement in a sentential complement & The bankers knew the \textbf{officer} \underline{smiles}/\underline{smile$^{*}$} \\
    Agreement across a prepositional phrase & The \textbf{farmer} near the parents \underline{smiles}/\underline{smile$^{*}$} \\
    Agreement across a subject relative clause & The \textbf{officers} that love the skater \underline{smile}/\underline{smiles$^{*}$} \\
    Short verb-phrase coordination & The \textbf{senator} smiles and \underline{laughs}/\underline{laugh$^{*}$} \\
    Long verb-phrase coordination & The \textbf{manager} writes in a journal every day and \\
    & ~\underline{likes}/\underline{like$^{*}$} to watch television shows \\
    Agreement across an objective relative clause & The \textbf{farmer} that the parents love \underline{swims}/\underline{swim$^{*}$} \\
    Agreement within an objective relative clause & The farmer that the \textbf{parents} \underline{love}/\underline{loves$^{*}$} swims
    \end{tabular}%
  \caption{Examples for the main syntactic constructions from the subject-verb-agreement stimuli from \citet{marvin-linzen-2018-targeted}. \textbf{Bold} indicates the subject-noun, and \underline{underlined} tokens indicate the grammatical/ungrammatical$^{*}$ verb.}
  \label{table:marvin_linzen_examples}%
\end{table*}%

\section{Token-level grammaticality}

We motivate the use of GED-probes by first reviewing previous literature involving grammaticality judgements and tests for subject-verb agreement, then discuss the advantages in tests for GED.

Boolean acceptability judgements have long been used as a primary behavioural measure to observe humans' grammatical knowledge \cite{chomsky1957syntactic, pater2019generative}, and have recently been employed in computational linguistics to evaluate grammatical knowledge in neural models. For example, \citet{warstadt2019neural} train classifiers to predict sentence-level Boolean acceptability judgements on example sentences from the linguistics literature. As each sentence is designed to demonstrate a particular grammatical construction, performance on the task is interpreted as a reflection of the implicit knowledge of the classifier. 

An alternative approach frames acceptability as a choice between minimal pairs of sentences -- one grammatical and another ungrammatical, where the difference between the two is typically one or two tokens. \citet{marvin-linzen-2018-targeted} evaluate linguistic knowledge by testing whether a language model assigns higher probability to a grammatical sentence relative to its minimally different ungrammatical counterpart. Similar to \citet{warstadt2019neural}, fine-grained grammatical knowledge is evaluated by controlling the evaluation stimuli, with the hypothesis that models must have implicit knowledge of the underlying grammatical concept to succeed.

% then introduce token-level grammaticality as a choice between verb-forms, but assuming detection.
Rather than evaluating sentence-level scores, \citet{linzen2016assessing} compare predicted probabilities assigned to target verbs in minimal pair sentences, where each sentence in a pair uses a different form of the verb. \citet{goldberg2019assessing} extends this to masked language models where he replaces a target verb with the [MASK] token and feeds the entire sentence to a \textsc{bert} model. A model is considered successful, and thereby has knowledge related to SVA, if it assigns higher probability to the correct form of the verb.

% Why is this a hard task? 
% Our work differs from the above in three important ways. First, we don't assume to know where the incorrect token is -- the probe is trained to detect errors for all tokens in a sentence, given each token's contextual representation. Second, instead of targeting information in the masked token, we investigate whether the model implicitly encodes SVA information in a token's contextual representation. Third, we test for knowledge of SVA without comparing to the counterpart token or sentence. If a model has encoded information for SVA error detection, 

Our work differs from the above in three important ways. First, we don't assume to know where the incorrect token is -- the probe is trained to detect errors for all tokens in a sentence, given each token's contextual representation. This is a more fine-grained evaluation compared to sentence-level judgements and tests whether probes know where the error is located. Second, instead of targeting information in the masked token, we investigate whether the model implicitly encodes SVA information in a token's contextual representation. Third, we test for knowledge of SVA without comparing to the counterpart token or sentence. We argue that if a model has knowledge of SVA, it should be able to detect SVA-errors without requiring a comparison. 

\section{Data} \label{section:data}

\subsection{Second language learner corpora} \label{section:learner_data}

Following previous work in grammatical error correction and detection, we use the Cambridge English Write \& Improve + LOCNESS corpus \citep{bryant2019bea} and the First Certificate in English \cite{yannakoudakis-etal-2011-new}, hereinafter \wifce.\footnote{Public data for W\&I and the FCE are available at: \url{https://www.cl.cam.ac.uk/research/nl/bea2019st\#data}}

The edit annotations in these corpora were pre-processed and standardised using the ERRANT annotation framework \cite{bryant2017automatic}. One advantage of this framework is that error types are modular, and consist of ``operation'' + ``main'' type tags. This provides us flexibility to target grammatical errors at different levels of granularity. E.g. all \texttt{NOUN} errors or only \texttt{R:NOUN} for replacement nouns. In addition, we can take advantage of the corrections provided with each edit annotation to control the number and variation of grammatical errors. Since we focus only on replacement subject-verb-agreement errors, \texttt{R:VERB:SVA}, we correct all other error types and keep only those sentences containing at least one grammatical error. This leaves 1936 sentences for training and 142 sentences for validation.

\subsection{Dataset of Wikipedia edits} \label{section:data_wiked}

As an alternative to the learner corpora, we additionally experiment with a corpus of automatically extracted edited sentences from native English Wikipedia edit histories (\wiked) \cite{wiked2014}. We use the clean and preprocessed version of English Wikipedia edits, consisting of \textasciitilde29 million sentences.\footnote{\url{https://github.com/snukky/wikiedits}} We follow the same procedure as above and retain only sentences with \texttt{R:VERB:SVA} errors by correcting all other error types, and keep only the sentences containing at least one error. This leaves \textasciitilde233K sentences, from which we sample five training sets each with 1936 sentences each to match the amount of sentences in the learner corpora after processing, and 5839 sentences for the validation set. Statistics for both corpora are given in \autoref{app:corpus_statistics}. We refer to the sampled training sets as \wikedsampled.

\subsection{Minimal-pair datasets} \label{section:marvin_linzen_data}

We use the manually constructed subject-verb agreement stimuli from \citet{marvin-linzen-2018-targeted} (\marvinlinzen) to evaluate the GED-probes -- this enables a more controlled evaluation compared to the naturally occurring sentences in \wifce{} and \wiked. The dataset consists of seven main syntactic constructions, shown in Table \ref{table:marvin_linzen_examples}. In addition to those shown in the table, sentences with multiple nouns (except for those testing VP coordination) include instances with two nouns and one acts as a distractor, potentially agreeing with the verb even though it is not the subject:

{
\small
\ex.\label{ex:prep_phrase_noun_number_agreement} 
  \a. The farmer near the parent smiles/smile$^{*}$.
  \b. The farmer near the parents smiles/smile$^{*}$.
  \c. The farmers near the parent smiles$^{*}$/smile.
  \d. The farmers near the parents smiles$^{*}$/smile.

}

In the above sentences, the verbs marked with an asterisk are ungrammatical. The dataset also expands on sentences testing agreement with object relative clauses: agreement is tested across and within the clause, using animate and inanimate main subjects, and with and without the \emph{that}-complementizer.

We process the \marvinlinzen{} minimal pairs to create token-level GED annotations, where the ungrammatical verbs are tagged as \texttt{R:VERB:SVA}.\footnote{In principle, any minimal-pair dataset can be converted to token-level annotations using ERRANT, but not all grammatical errors map cleanly to ERRANT categories. For example, replacement pronouns (\texttt{R:PRON}) includes reflexive anaphor gender- and number- agreement errors.} We include all of the ungrammatical and grammatical sentences for evaluation -- to be successful, the GED-probe should recognise when the agreement is correct and label all tokens as grammatical.\footnote{While the evaluation stimuli consists of minimal pairs, the training data does not.} Finally, we capitalise the first word in each sentence and add a full stop if one doesn't already exist. \autoref{app:marvin_linzen_dataset_statistics} contains details about the processed dataset.

\section{Experiment 1: Per Layer Probes}
\label{section:per_layer_probes}

We first investigate whether models encode SVA-errors by examining probing performance at each layer; we want to test whether models encode this information in the final layer, the token representation, but also how this information develops across the layers. We then break-down performance by syntactic construction to better understand how the SVA-error encoding generalizes. Finally, we carry out a follow-up experiment to investigate the impact of training data size and verb frequency. 

\subsection{Experimental setup}

For each model, we extract contextual representations for every token in a sentence for every one of the twelve layers in the model. We then train a linear probe \cite{tenney2019you, hupkes2018visualisation, Liu2019-lg} per layer to predict whether each token-level contextual representation is ungrammatical. We train two versions of each probe: one trained using \wifce, and the other using \wikedsampled. Every probe is evaluated on the \marvinlinzen{} stimuli 

Since the probe is trained to detect \texttt{R:VERB:SVA} errors, high probing performance would indicate the probe has learned to extract features to identify subject-verb agreement errors from the contextual representations of the verbs. This implicitly includes sub-tasks to identify the verb, the subject noun, the number of both the verb and noun, and that their number disagrees. Furthermore, as this is a token labelling probe, high probing performance would indicate the pre-trained model has encoded the relevant features in the contextual representations of the verbs.

We evaluate probes using F$_{1}$ on the evaluation stimuli from \marvinlinzen{} containing an equal number of grammatical and ungrammatical sentences.\footnote{This departs from F$_{0.5}$ used in the GED literature, which was motivated from educational applications where high precision is preferred over recall because false-positives can be more harmful for language learners compared to false-negatives.} We compare probes to a \textsc{verb-only} baseline which incorrectly tags all verbs as ungrammatical. The number of verbs per sentence varies across syntactic constructions; constructions with one verb have an equal number of grammatical and ungrammatical verbs, and therefore have a baseline score of 0.67. Constructions with two and three verbs have scores of 0.40 and 0.30, respectively. Evaluating the baseline over all constructions yields a score of 0.43.

We evaluate five pre-trained models: \textsc{bert-base-cased}, \textsc{gpt-2} (small), \textsc{roberta-base}, \textsc{xlnet} with both uni-directional (\textsc{xlnet-uni}) and bi-directional \textsc{xlnet-bi} decoding, and \textsc{electra-base} (discriminator). As all five models use sub-word tokenisation, we follow \citet{Liu2019-lg} and use the last sub-word unit for token classification. We train the probes for 50 epochs with a patience of 10 epochs for early stopping based on in-domain validation sets.

Four of the five models were pre-trained with a language modelling objective: either masked language modelling (MLM) or autoregressive language modelling (ALM).  Whereas \textsc{electra} is the exception -- the \emph{replaced token detection} training objective is somewhat aligned with GED and therefore we may expect representations to encode grammatically discriminative information.\footnote{The \textsc{electra} discriminator model is trained to detect substituted tokens in a grammatical sentence, where an original token is substituted with a plausible alternative from a masked language model.} Indeed, \citet{yuan2021multiclass} find \textsc{electra} outperforms \textsc{bert} when fine-tuned for binary GED targeting a wide range of error-types. \textsc{bert} and \textsc{roberta} also detect replaced tokens during training, but only on 1.5\% of tokens.

\begin{figure*}
     \centering
     \begin{subfigure}[b]{0.49\textwidth}
         \centering
         \includegraphics[width=\textwidth]{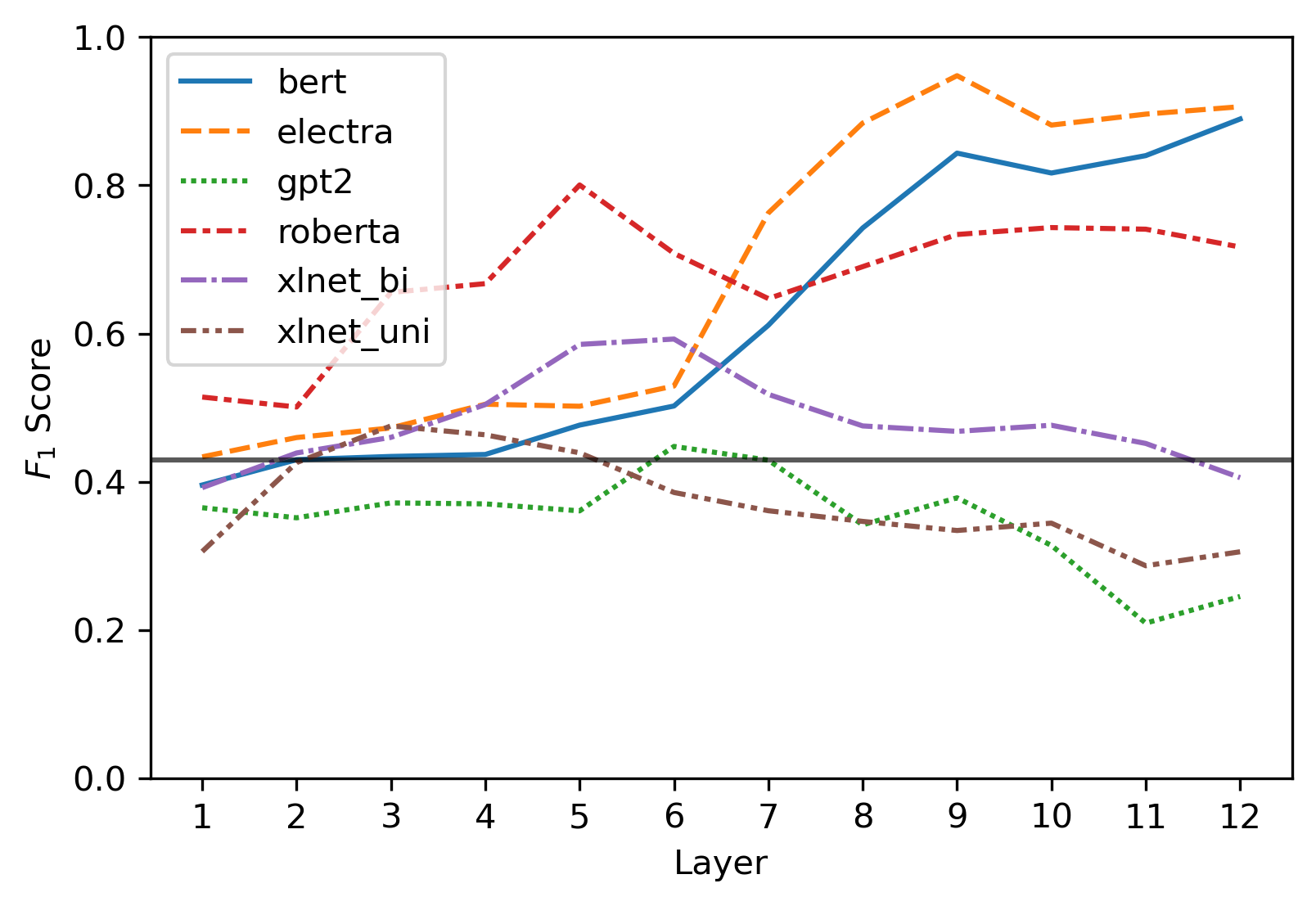}
         \caption{\wifce}
         \label{fig:wibea_marvinlinzen_overview}
     \end{subfigure}
     \hfill
     \begin{subfigure}[b]{0.49\textwidth}
         \centering
         \includegraphics[width=\textwidth]{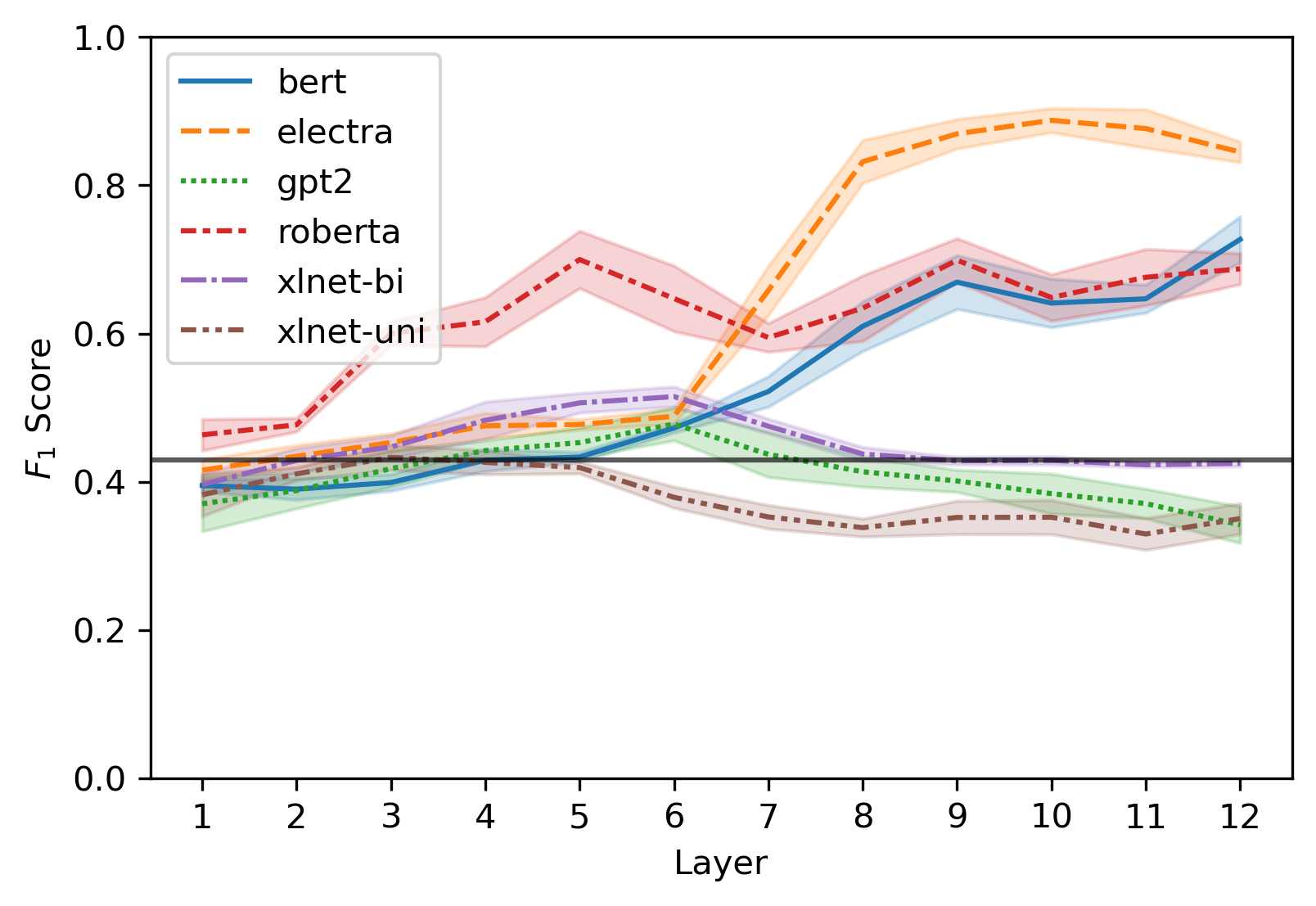}
         \caption{\wikedsampled}
         \label{fig:wiked_marvinlinzen_overview}
     \end{subfigure}
    \caption{F$_{1}$ scores for probes trained on contextual representations at different layers from \textsc{bert}, \textsc{electra}, \textsc{roberta}, \textsc{xlnet} with bidirectional decoding (\textsc{xlnet-bi}), \textsc{xlnet} with unidirectional decoding (\textsc{xlnet-uni}), and \textsc{gpt-2}. \ref{fig:wibea_marvinlinzen_overview} and \ref{fig:wiked_marvinlinzen_overview} show results for probes trained on \wifce{} and the \wikedsampled{} training sets respectively. The \textsc{verb-only} baseline scores are illustrated using grey horizontal lines. \ref{fig:wiked_marvinlinzen_overview} shows the mean and standard deviation across the five training sets (\cref{section:data_wiked}). All probes are evaluated on the \marvinlinzen{} stimuli (\cref{section:marvin_linzen_data}).}
    
    \label{fig:marvinlinzen_model_overview}
\end{figure*}

\begin{table}[t!]
    \footnotesize
    \centering
    \begin{tabular}{lrrrrr}
    \toprule
    & \multicolumn{2}{c}{\emph{\wifce}} & \multicolumn{2}{c}{\emph{\wikedsampled}} \\[4pt]
    Model & Layer & F$_{1}$ & Layer & F$_{1}$ \\
    \midrule
    \textsc{electra} &           9 &      0.95 &          10 &     0.89$_{\pm0.02}$ \\
    \textsc{bert} &          12 &      0.89 &          12 &     0.73$_{\pm0.03}$ \\
    \textsc{roberta} &           5 &      0.80 &           5 &     0.70$_{\pm0.04}$ \\
    \textsc{xlnet-bi} &         6 &      0.59 &           6 &     0.51$_{\pm0.01}$ \\
    \textsc{xlnet-uni} &          3 &      0.48 &           3 &     0.43$_{\pm0.02}$ \\
    \textsc{gpt-2} &           6 &      0.45 &           6 &     0.48$_{\pm0.02}$ \\
    \midrule
    \textsc{verb-only} & - & 0.43 & - & 0.43 \\
    \bottomrule
    \end{tabular}
    \caption{Top F$_{1}$ scores on the \marvinlinzen{} evaluation set, for probes trained on either \wifce{} or \wikedsampled{}. Scores for probes trained on \wikedsampled{} are reported as the \emph{mean} $\pm$1 standard deviation over the five sampled training sets.}
    \label{table:max_f1_scores}
    
\end{table}%

\subsection{Results}

% intro statement
Figure \ref{fig:marvinlinzen_model_overview} shows F$_{1}$ scores for probes trained on \wifce{} and \wikedsampled{}, and evaluated on the \marvinlinzen{} stimuli.\footnote{We additionally evaluate probes against the other error types in the W\&I dataset and verify that probes only detect SVA errors. Probes trained on either \textsc{bert} or \textsc{electra} obtain mean scores of 0.04 ($\sigma$=0.04), verifying that information extracted by the probe is isolated to subject-verb agreement errors.} For the latter, we plot the mean and standard deviation evaluated over the five sampled training sets. We illustrate the \textsc{verb-only} baseline score with a grey horizontal line. Table \ref{table:max_f1_scores} shows layers which obtained the top F$_{1}$ score per model.

% results for electra - it does the best, we need to mention it
The figure and table shows \textsc{electra} encodes the most salient information for SVA error detection, with probes obtaining maximum scores of 0.95 and 0.89 ($\sigma$=0.02) when trained on \wifce{} and \wikedsampled{}, respectively. Though this may not be surprising given the \emph{replaced token detection} pre-training objective, it illustrates that probes trained on representations from a model capable of SVA error detection can obtain high performance using both training sets.

% difference between MLMs versus ALMs
We observe a divergence in performance between MLM-probes and ALM-probes; the MLM-probes tend to perform better, obtaining maximum scores between 0.70 and 0.89 F$_{1}$, while the ALM-probes don't score above 0.59 on either training set. In fact, probes trained on representations from \textsc{gpt-2} and \textsc{xlnet-uni} often perform worse than the \textsc{verb-only} baseline at 0.43 and don't score above 0.50 F$_{1}$. These results imply that \textsc{gpt-2} and \textsc{xlnet-uni} representations do not linearly encode enough information to differentiate between grammatical and ungrammatical verbs in SVA.

% LMs do well on SVA-error detection, though not trained for it
The MLM-ALM performance gap could be due to the language model directionality: the two uni-directional models (\textsc{gpt-2} and \textsc{xlnet-uni}) do perform the worst, but this fails to account for the performance of \textsc{xlnet-bi} -- a bi-directional language model which does not perform much better. It may be that the MLM training objective helps to imbue contextual representations with information useful for detecting SVA errors, but we cannot discount the inclusion of the \textit{replaced token detection} objective, even though it is rarely included. Finally, we note key differences in the pre-training data used by the models: \textsc{bert} and \textsc{electra} use the BooksCorpus and English Wikipedia, \textsc{gpt-2} uses web-scraped data, and \textsc{xlnet} and \textsc{roberta} use a combination of BooksCorpus, Wikipedia, and web-scraped data.

% The divergence between models could be due to different inductive biases imposed by uni- and bi-directional language models, given that the two uni-directional models (\textsc{gpt-2} and \textsc{xlnet-uni}) perform the worst, but this fails to account for the performance of \textsc{xlnet-bi} -- a bi-directional language model which does not perform much better. An alternative explanation could be that the MLM training objective is responsible for imbuing contextual representations with information useful for detecting SVA errors. Further still, we note key differences in the data used for pre-training the models: \textsc{bert} and \textsc{electra} use the BooksCorpus and English Wikipedia, \textsc{gpt-2} uses web-scraped data, and \textsc{xlnet} and \textsc{roberta} use a combination of BooksCorpus, Wikipedia, and web-scraped data. \textsc{roberta} seems to fall into a middle-category, indicating that both the training objective and data are contributing factors.

% summary results across layers for electra, bert, and roberta.
When we examine performance across layers we see that \textsc{electra}- and \textsc{bert}-probes follow a similar trajectory, with performance on par with the baseline from layers 1-5 and higher performance only in layers 8-12. SVA-error information is highest in the final layer for \textsc{bert} (the token representation) while layers 9 and 10 seem to encode the most useful information for \textsc{electra}. In contrast, probes trained with representations from \textsc{roberta} peak at layer 5 but have mostly consistent performance until 12, apart from a drop in layer 7. 

The results for \textsc{bert} support those from \citet{jawahar:hal-02131630}, where they find probes encode elements of syntax in the middle to late layers. \citet{Liu2019-lg} also find that later layers obtain the best performance for a general GED probe, though they experiment on a single dataset (FCE \cite{yannakoudakis-etal-2011-new}) and include all grammatical error types. Recent results from \citet{lasri2022probing} show that removing \emph{number} information from nouns at different layers has a detrimental effect on the number-agreement task up until layer 8. They hypothesise that some transfer of noun-\emph{number} takes place in the previous layers. Our results seem to support this: low probing performance in layers 1-5 (when the noun-\emph{number} has yet to be transferred to the verb), and high probing performance in layers 8-12 after transfer has taken place. Interestingly, performance for \textsc{electra} takes the same shape, suggesting that the pre-training objective (\emph{replace token detection} versus MLM) may not have an important role in how models encode SVA information, especially given that \textsc{roberta}, trained using masked-language modelling, displays a different pattern across layers. Finally, if GED-probe-performance can be taken as a proxy for noun-\emph{number} transfer, then results for \textsc{roberta}-probes suggest that noun-\emph{number} information is transferred to the target verb earlier, possibly due to the more robust optimization.

% difference in performance between training sets (WI+FCE versus WIKI)
Turning to probe performance across training sets, we find that probes trained on \wifce{} consistently perform better than those trained on \wikedsampled{}, for all models tested apart from \textsc{gpt-2}. This could be due to a domain mismatch, where learner writing may be more similar to the \marvinlinzen{} stimuli than data from \wikedsampled{}. Though, at the very least this indicates that information pertaining to SVA errors is not consistently encoded.

% -- if models had robust and consistent knowledge of SVA then they should be able to consistently detect when there are SVA errors, though we concede that our investigation is limited to linear encoding.

% Additionally, we find performance on the \marvinlinzen{} stimuli is generally lower for probes trained using \wikedsampled{} compared to \wifce{}, for all models apart from \textsc{gpt-2}. 

% The fluctuation in results across training set and across syntactic construction show that information related to broken SVA is not encoded consistently across verbs. Previous results using targeted syntactic tests show models do encode a large amount of SVA information sufficient to distinguish the correct from incorrect verb form. And various probing experiments show that number can be predicted using linear probes. The results suggest that although models may have some knowledge of hierarchical syntax (related to SVA), they are not able to consistently identify broken SVA.

Previous work finds some evidence that \textsc{bert}'s representations encode knowledge of SVA \cite{goldberg2019assessing, jawahar:hal-02131630, newman2021refining, lasri2022does} but that this knowledge is based on heuristics rather than robust SVA rule learning \cite{chaves2021look, mccoy2019right}. Our results indicate that information encoded in contextual representations extends to the detection of SVA errors in \textsc{electra}, \textsc{bert}, and to a lesser extent, \textsc{roberta}. However, we find the encoding is not robust across domains, supporting the heuristic-learning claim. These results illustrate the importance of utilising training and evaluation datasets from disparate domains to evaluate probes.

\begin{figure*}
    \centering
    \includegraphics[width=0.9\textwidth]{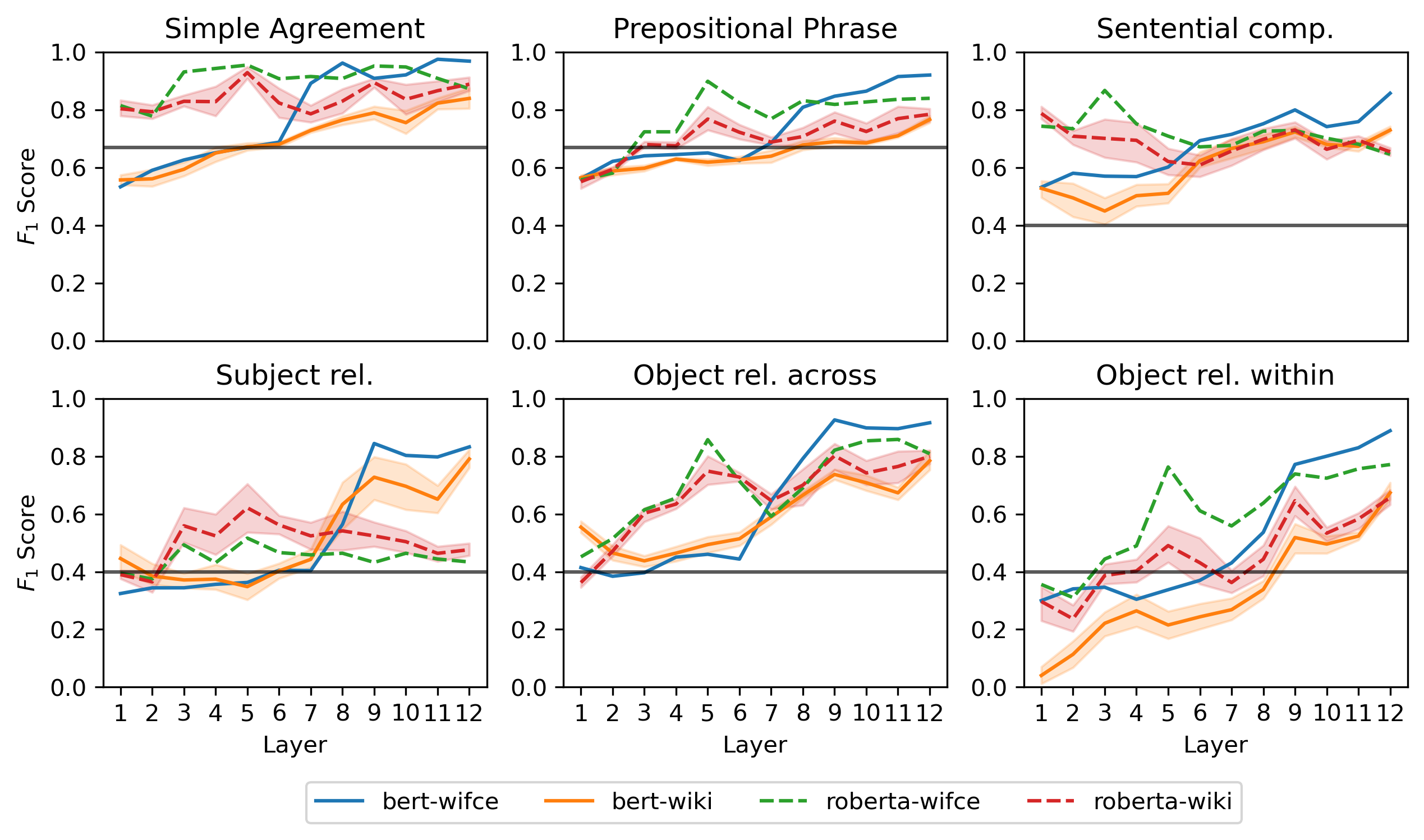}
    \caption{F$_{1}$ scores for probes trained on contextual representations from \textsc{bert} and \textsc{roberta}, using both the \wifce{} and \wikedsampled{} training sets. The probes are evaluated on \marvinlinzen{} stimuli. The \textsc{verb-only} baseline is illustrated using grey horizontal lines.}
    \label{fig:results_per_syntactic_construction}
\end{figure*}

\subsection{Results per syntactic construction}
\label{section:results_per_construction}

% To discuss: which syntactic constructions cause the most difference between probes?
% for probes trained on \textsc{bert} and \textsc{roberta} representations 
% dive deeper into the results by breaking

We break down the performance of probes for each syntactic construction in the \marvinlinzen{} dataset. Figure \ref{fig:results_per_syntactic_construction} illustrates F$_{1}$ scores for probes trained on \wifce{} and \wikedsampled{} using representations from each layer of \textsc{bert} and \textsc{roberta}.\footnote{We select \textsc{bert} and \textsc{roberta} because they are both MLMs, whereas the \textsc{electra}-discriminator is trained using a \emph{replaced token detection} objective.} The \textsc{verb-only} baseline is shown as grey horizontal lines. For brevity, we present results on sentences with simple agreement, sentential complements, prepositional phrases, subject relative clauses, and object relative clauses (agreement within and across the clause). Results for the other models and syntactic constructions are included in \autoref{app:probe_results_syn_const}. 

% trends across layers
The trends across layers observed in Figure \ref{fig:marvinlinzen_model_overview} are generally consistent within syntactic constructions: probes trained on \textsc{bert} representations improve in the later layers, while layers 5 and 8-12 seem to be the most salient for probes trained on \textsc{roberta}.

% bert versus roberta - models do well on:
Probes for both models detect SVA errors in the simple agreement constructions -- with only the \textsc{bert}-probe trained on \wikedsampled{} scoring less that 0.90 F$_{1}$. We find \textsc{bert}-probes perform well for most constructions, especially in layer 12, but performance for \textsc{roberta}-probes drops for sentences with subject-relative clauses, prepositional phrases, and object-relative clasues (agreement within the clause), particularly when trained on \wikedsampled{} data. This suggests that the token representation from \textsc{bert} models, potentially used in downstream tasks, already encodes a lot of information related to SVA before any fine-tuning.

% wifce versus wiki
When comparing performance between probes across training sets, we observe a noticeable difference in performance between \textsc{bert}-probes trained on \wifce{} versus those trained on \wikedsampled, most evidently in five out of the six syntactic constructions shown. On the other hand, probes trained on \textsc{roberta} don't always display a performance gap -- performance is more comparable between probes on sentences testing simple agreement, sentential complements, and agreement across object relative clauses. For sentences with subject relative clauses we find the probes trained on \wikedsampled{} outperform those trained on \wifce{}. These results may indicate that information pertaining to the detection of SVA errors is more robustly encoded in \textsc{roberta} than \textsc{bert} -- that is, the information is more invariant to the choice of probe training set.

%%%%%%

\begin{figure*}
    \centering
    \includegraphics[width=\textwidth]{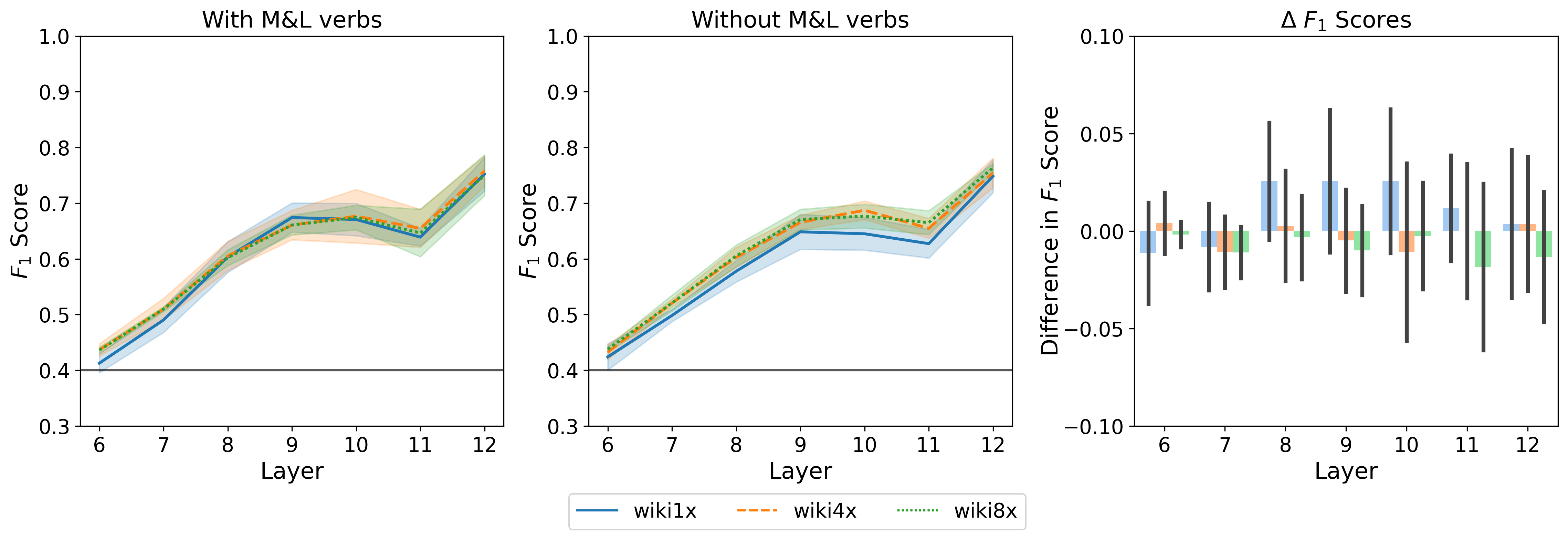}
    \caption{F$_{1}$ scores for probes trained on \textsc{bert} representations, with varying amounts of data (left and centre); 1x=1936, 4x=7744, 8x=15488 sentences. The plots show the mean and standard deviation across five training sets (\cref{section:exp_2_generalisation}). On the left, probes are trained on data including \marvinlinzen{} verbs, while the centre shows scores for probes trained on data without \marvinlinzen{} verbs (apart from ``is'' and ``are'', as described in \cref{section:exp_2_generalisation}). The right-hand plot shows the mean and standard deviation of differences in F$_{1}$ scores between probes trained on the two types of dataset: with and without \marvinlinzen{} verbs. The \textsc{verb-only} baseline is shown as a grey horizontal line.}
    \label{fig:hold_out_exp}
\end{figure*}

\section{Experiment 2: Generalisation to unseen verbs}
\label{section:exp_2_generalisation}

In our first experiment, we observe that although the MLMs encode more information for SVA error detection compared to the ALMs, the information does not always generalize across domains or syntactic constructions. We carry out a follow-up experiment to investigate whether the information generalizes across verbs using probes trained on \textsc{bert} representations and \wiked{} data. We focus on layers 6 to 12 as these were the layers where performance was above the baseline. There are 13 target verbs in the \marvinlinzen{} stimuli, of which ``to be'' is the most frequent with 946 occurrences in \wikedsampled{}. The remaining verbs appear very infrequently -- for example, eight verbs have frequencies less than 30. To test generalization across verbs we remove all sentences from the training and development sets which contain a verb from the \marvinlinzen{} stimuli except for ``to be''. We then re-sample sentences from the full \wiked{} data to maintain 1936 sentences as in the first experiment. Due to the infrequency of many verbs and to ensure a more thorough evaluation, we also increase the training set size by 4- and 8-times to yield training sets with 7744 and 15488 sentences, respectively. We refer to the three sizes as small, medium, and large. This results in paired training sets: for each training set size, there is one set ``with \marvinlinzen{} verbs'' and a set ``without \marvinlinzen{} verbs''. We sample each training set five times and report the mean and standard deviation over the samples. For example, we sample five training sets with 7744 sentences ``with \marvinlinzen{} verbs'', and another five ``without \marvinlinzen{} verbs''. Since we are only interested in the performance of 12 verbs, we modify the evaluation stimuli to remove a) sentences containing only ``to be'' verbs, and for sentences with multiple verbs we remove the ``to be'' token from evaluation. For example, in the sentence ``The movie the security guards like is good'', we remove ``is'' from evaluation.

Figure \ref{fig:hold_out_exp} illustrates the F$_{1}$ scores: the left and centre plots show results for probes trained on data with and without \marvinlinzen{} verbs, respectively. The plot on the right presents the mean and standard deviation for the pairwise differences between probes trained on datasets of the same size.
The right-hand plot shows that for the smallest training set size we observe a slight benefit when including the \marvinlinzen{} verbs, but this is limited to \textasciitilde0.05 F$_{1}$ and restricted to layers 8-10. We generally observe no difference in performance for the medium and large training sets, indicating that SVA-error information does generalize across verbs. Furthermore, we observe no difference when comparing results across training set sizes in the left and central plots, except for probes trained on the small training set without \marvinlinzen{} verbs. These results indicate that SVA-error information is linearly accessible and generalizable across across the verbs we test, even when probes are trained with limited data. Future work may expand the investigation to cover more verbs, though we expect performance to deteriorate as verbs become infrequent in the pre-training data \cite{wei2021frequency}.

\section{Discussion}

Our experiments test whether information for SVA errors is implicitly encoded in the contextual representations of verbs, but they don't provide any indication as to \emph{how} the information is encoded: is grammaticality encoded atomically or compositionally? Furthermore, we note that selecting the ``erroneous token'' can be an ambiguous choice between the noun and the verb, for example in ``The authors laughs''. Yet, the probes we evaluate never tag the nouns. This could indicate that a) the probes learn to only tag verbs, and/or b) that SVA-grammaticality is disparately encoded between nouns and verbs. A compositional account of grammatical encoding is a plausible explanation given the results provided in \citet{lasri2022probing} -- that nouns and verbs have different encodings for \emph{number}. We plan to investigate how grammaticality is encoded in future work, both in pre-trained language models as well as models trained specifically for GED.

\vspace{-0.5em}

\section{Conclusion}

We analyse whether pre-trained transformer-based language models implicitly encode knowledge of SVA errors using GED probes. We carry out a thorough evaluation on five models, using two public training sets from different domains, and evaluate on a manually constructed evaluation set. This enables us to get a more complete and reliable picture of a models' performance.

Grammatical error detection is a challenging and linguistically aligned task to assess the knowledge of neural language models; we show that GED-probes can be used as a complementary analysis tool to evaluate a models' linguistic capabilities.

Our results show that \textsc{electra}, \textsc{bert}, and \textsc{roberta} encode information for SVA-error detection, but \textsc{gpt-2} and \textsc{xlnet} do not. For \textsc{bert} and \textsc{roberta}, we find that the SVA-error encoding is not robust across all syntactic constructions or training set domains, though we do find some evidence that the encoding generalizes across verbs for \textsc{bert}. Furthermore, a layer-wise analysis reveals the final layers in \textsc{electra} and \textsc{bert} are the most salient for SVA-error detection.

\section*{Acknowledgements}

We thank Guy Aglionby, Rami Aly, Paula Czarnowska, and Tiago Pimentel for helpful discussions and feedback on early drafts of this work. We also thank the anonymous reviewers for their helpful feedback. This paper reports on research supported by Cambridge University Press \& Assessment. We thank the NVIDIA Corporation for the donation of the Titan X Pascal GPU used in this research.

\bibliography{anthology,custom}
\bibliographystyle{acl_natbib}

\newpage

\appendix

\onecolumn

\section{Corpus statistics}
\label{app:corpus_statistics}

\begin{table}[h!]
\centering
\begin{tabular}{llrrr}
\hline
& \textbf{Corpus} & \textbf{\# sentences} & \textbf{$\mu$ sent. length ($\sigma$)} & \textbf{$\mu$ errors per sentence ($\sigma$)} \\
\hline
\multirow{3}{4em}{Original} & FCE-train & 28K & 16 (11) & 1.9 (2.5) \\
& W\&I-train & 34K & 18 (12) & 2.0 (2.8) \\
& BEA-dev & 4K & 20 (12) & 1.9 (2.8)  \\
& WikEd (total) & 28M & 22 (12) & 1.6 (1.6) \\
\hline
\multirow{3}{4em}{Processed} & FCE-train & 626 & 25 (14) & 1.1 (0.3) \\
& W\&I-train & 1310 & 27 (21) & 1.1 (0.3) \\
& BEA-dev & 142 & 26 (19) & 1.1 (0.3) \\
& WikEd-train & 1936 & 24 (12) & 1.0 (0.2) \\  
& WikEd-dev & 5839 & 23 (11) & 1.0 (0.2) \\
\hline
\end{tabular}%
\caption{Corpus statistics.}
\label{table:probe-training-corpus-stats}
\end{table}

\section{Marvin \& Linzen statistics}
\label{app:marvin_linzen_dataset_statistics}

Statistics per construction.

\begin{table}[h!]
\centering
\begin{tabular}{lrrr}\toprule
Construction & \textbf{\# sentences} & \textbf{$\mu$ sent. length ($\sigma$)} \\\midrule
Simple agr. &280 &4.57 (0.49) \\
In sent. comp. &3360 &7.57 (0.49) \\
Across prep. &44800 &8.85 (1.17) \\
Across subj. rel. &22400 &8.77 (0.64) \\
Short VP coord &1680 &7.14 (0.64) \\
Long VP coord &800 &14.40 (0.49) \\
Across obj. rel. &44800 &9.18 (0.86) \\
Across obj. rel. (no comp) &44800 &8.18 (0.86) \\
Within obj. rel. &44800 &9.18 (0.86) \\
Within obj. rel. (no comp) &44800 &8.18 (0.86) \\
\bottomrule
\end{tabular}
\caption{Details for the evaluation stimuli from \cite{marvin-linzen-2018-targeted}.}
\label{table:marvin-linzen-stats}
\end{table}

\newpage

\section{Results for probes trained on \wifce{}}
\label{app:probe_results_wifce}

\begin{table}[h]
\footnotesize
\centering
\begin{tabular}{lRRRRRRRRRRRR}
\toprule
    Model &1 &2 &3 &4 &5 &6 &7 &8 &9 &10 &11 &12 \\
    \midrule
    \textsc{bert} &0.40 &0.43 &0.43 &0.44 &0.48 &0.50 &0.61 &0.74 &0.84 &0.82 &0.84 &0.89 \\
    \textsc{electra} &0.43 &0.46 &0.47 &0.50 &0.50 &0.53 &0.76 &0.88 &0.95 &0.88 &0.90 &0.91 \\
    \textsc{roberta} &0.51 &0.50 &0.66 &0.67 &0.80 &0.71 &0.65 &0.69 &0.73 &0.74 &0.74 &0.72 \\
    \textsc{gpt-2} &0.37 &0.35 &0.37 &0.37 &0.36 &0.45 &0.43 &0.34 &0.38 &0.31 &0.21 &0.25 \\
    \textsc{xlnet-bi} &0.39 &0.44 &0.46 &0.50 &0.59 &0.59 &0.52 &0.48 &0.47 &0.48 &0.45 &0.41 \\
    \textsc{xlnet-uni} &0.31 &0.43 &0.48 &0.46 &0.44 &0.39 &0.36 &0.35 &0.33 &0.34 &0.29 &0.31 \\
    \bottomrule
\end{tabular}
\caption{F$_{1}$ scores for probes trained on contextual representations at different layers from \textsc{bert}, \textsc{electra}, \textsc{roberta}, \textsc{xlnet} with bidirectional decoding, \textsc{xlnet} with unidirectional decoding, and \textsc{gpt-2}. Probes were trained on learner data described in \cref{section:learner_data}, and evaluated on the \citet{marvin-linzen-2018-targeted} stimuli (\cref{section:marvin_linzen_data}).}
\end{table}

\section{Results for probes trained on \wikedsampled{}}
\label{app:probe_results_wiked}

\begin{table}[h]
\footnotesize
\centering
\begin{tabular}{lRRRRRRRRRRRR}
\toprule
    Model &1 &2 &3 &4 &5 &6 &7 &8 &9 &10 &11 &12 \\
    \midrule
    \textsc{bert} &0.39 &0.39 &0.40 &0.43 &0.43 &0.47 &0.53 &0.62 &0.68 &0.65 &0.65 &0.73 \\
    \textsc{electra} &0.42 &0.43 &0.45 &0.48 &0.48 &0.49 &0.66 &0.83 &0.87 &0.89 &0.88 &0.84 \\
    \textsc{roberta} &0.46 &0.48 &0.60 &0.62 &0.70 &0.65 &0.59 &0.63 &0.70 &0.65 &0.68 &0.69 \\
    \textsc{gpt-2} &0.37 &0.39 &0.42 &0.44 &0.45 &0.48 &0.44 &0.41 &0.40 &0.38 &0.37 &0.34 \\
    \textsc{xlnet-uni} &0.38 &0.41 &0.43 &0.43 &0.42 &0.38 &0.35 &0.34 &0.35 &0.35 &0.33 &0.35 \\
    \textsc{xlnet-bi} &0.40 &0.43 &0.45 &0.48 &0.51 &0.51 &0.48 &0.44 &0.43 &0.43 &0.42 &0.43 \\
    \bottomrule
\end{tabular}
\caption{F$_{1}$ scores for probes trained on contextual representations at different layers from \textsc{bert}, \textsc{electra}, \textsc{roberta}, \textsc{xlnet} with bidirectional decoding, \textsc{xlnet} with unidirectional decoding, and \textsc{gpt-2}. Probes were trained on wikipedia data described in \cref{section:data_wiked}, and evaluated on the \citet{marvin-linzen-2018-targeted} stimuli (\cref{section:marvin_linzen_data}).}
\end{table}

\newpage

\section{Results across syntactic constructions}
\label{app:probe_results_syn_const}

\begin{figure}[h]
    \centering
    \includegraphics[width=0.9\textwidth]{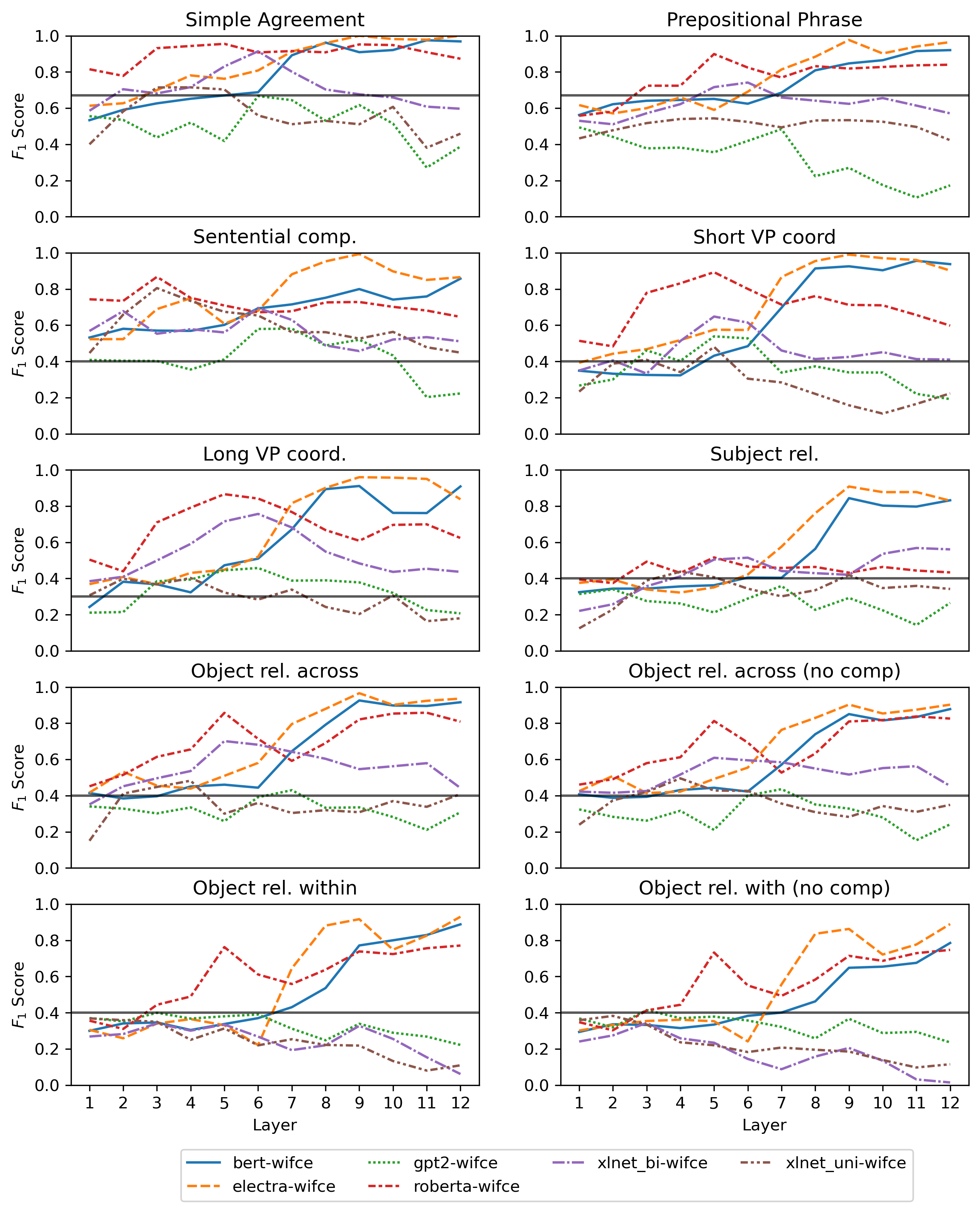}
    \caption{F$_{1}$ scores for probes trained the \wifce{} and training set. The probes are evaluated on \marvinlinzen{} stimuli. The \textsc{verb-only} baseline is illustrated using grey horizontal lines.}
    \label{fig:wifce_results_per_all_syntactic_construction}
\end{figure}

\begin{figure}[h]
    \centering
    \includegraphics[width=0.9\textwidth]{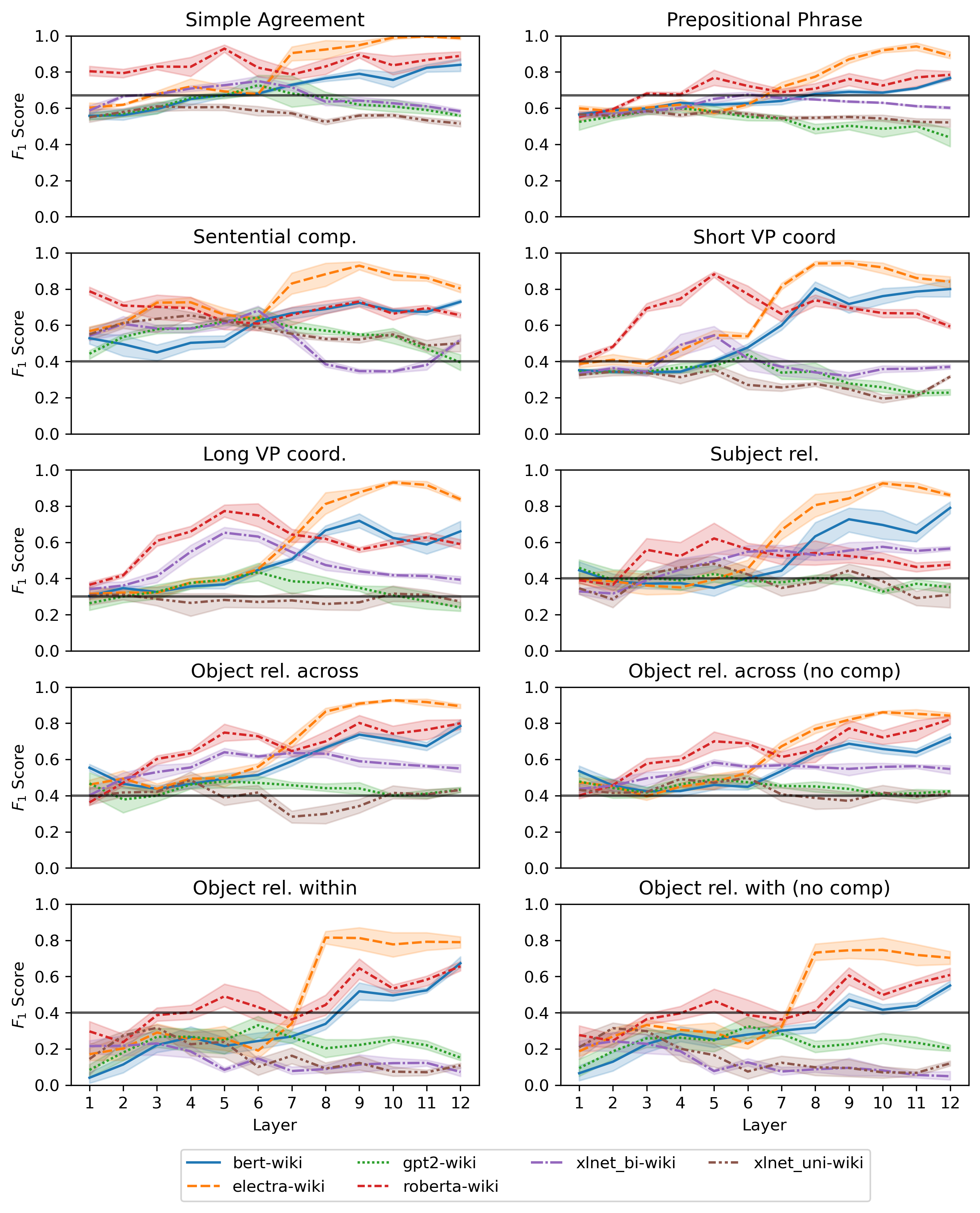}
    \caption{F$_{1}$ scores for probes trained the \wiked{} and training set. The probes are evaluated on \marvinlinzen{} stimuli. The \textsc{verb-only} baseline is illustrated using grey horizontal lines.}
    \label{fig:wiki_results_per_all_syntactic_construction}
\end{figure}

\end{document}